\documentclass{article} 
\usepackage{iclr2027_conference,times}

\usepackage{amsmath,amsfonts,bm}

\def\eqref#1{equation~\ref{#1}}

\def\1{\bm{1}}

\DeclareMathAlphabet{\mathsfit}{\encodingdefault}{\sfdefault}{m}{sl}
\SetMathAlphabet{\mathsfit}{bold}{\encodingdefault}{\sfdefault}{bx}{n}

\usepackage{hyperref}
\usepackage{url}

\usepackage[utf8]{inputenc} 
\usepackage[T1]{fontenc}    
\usepackage{hyperref}       
\usepackage{url}            
\usepackage{booktabs}       
\usepackage{amsfonts}       
\usepackage{nicefrac}       
\usepackage{microtype}      
\usepackage{xcolor}         
\usepackage{comment}
\usepackage{graphicx}
\usepackage{subcaption}
\usepackage[export]{adjustbox}
\usepackage{amsmath}
\usepackage{enumitem}      

\usepackage[disable,textsize=tiny]{todonotes}

\title{Efficient Mixture-of-Experts with \\ Speculative Decoding via Expert Coactivation}

\author{Kumari Nishu, Han-Byul Kim, Santosh Chilkunda, Maxwell Horton\thanks{Contributed when employed by Apple.}\,, \AND
Arnav Kundu\footnotemark[1]\,, Mohammad Samragh, Lauren Hannah, Mohammad Sekhavat, \AND
Nikhil Bhendawade, Manuel Ciosici, Iman Mirzadeh, Keivan Alizadeh Vahid, \AND
David Harrison, Irina Belousova, Mehrdad Farajtabar \& Minsik Cho \\
Apple
}

\iclrfinalcopy
\begin{document}

\maketitle

\begin{abstract}
Mixture-of-Experts (MoE) models are increasingly deployed alongside Speculative Decoding (SD) to accelerate inference, but combining the two is challenging. SD improves the inference speed of dense models by verifying groups of tokens in parallel. However, the inference speedup for SD with MoEs depends heavily on the number of tokens being verified. Using more verification tokens results in more experts being transferred from DRAM to the Neural Processing Unit (NPU), which increases the memory transfer cost. This negatively impacts model runtime, as memory transfer is typically the bottleneck in inference. In this work, we investigate the impact of MoE router design during training on the speed of MoEs with SD. We find that routers with high degrees of expert coactivation result in much faster runtimes, mitigating the impact of using more verification tokens. Motivated by this observation, we assess the impact of various router design choices on expert coactivation and runtime using billion-parameter transformer models. We find that combining a global load-balancing loss, shared experts, a consistency loss, and an autoregressive expert selection mechanism during training results in significantly stronger expert coactivation. This increased coactivation translates into higher overall runtime throughput: our exploration yields a model that improves throughput by $21\%$ over MoE baselines, while maintaining on-par accuracy with the baseline MoE.
\end{abstract}

\begin{figure}[h!]
\begin{subfigure}[t]{0.24\textwidth}
    \centering
    \adjincludegraphics[page=2,trim={0 0 {.75\width} {0.5\height}},clip=true,width=\textwidth]{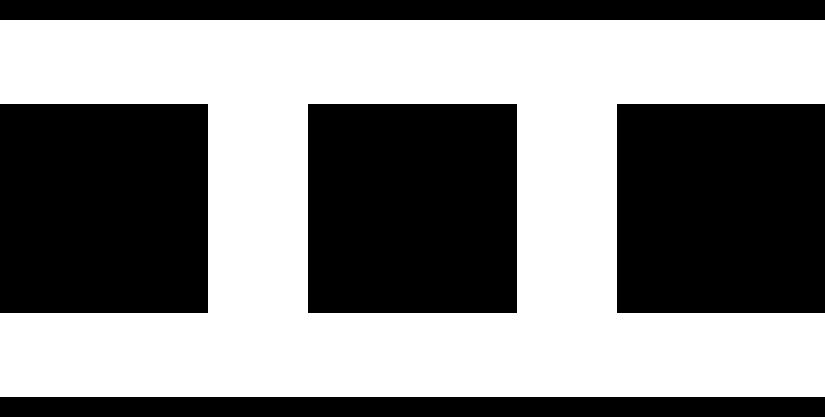}
    \caption{MoE}
    \label{fig:overview-moe}
\end{subfigure}
\begin{subfigure}[t]{0.24\textwidth}
    \centering
    \adjincludegraphics[page=3,trim={0 0 {.75\width} {0.5\height}},clip=true,width=\textwidth]{figures/teaser.pdf}
    \caption{Our MoE}
    \label{fig:overview-ours}
\end{subfigure}
\begin{subfigure}[t]{0.24\textwidth}
    \centering
    \includegraphics[width=\textwidth]{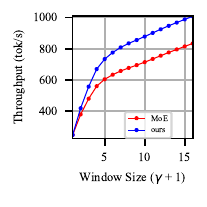}
    \caption{Throughput}
    \label{fig:overview-throughput-vs-gamma}
\end{subfigure}
\begin{subfigure}[t]{0.24\textwidth}
    \centering
    \includegraphics[width=\textwidth]{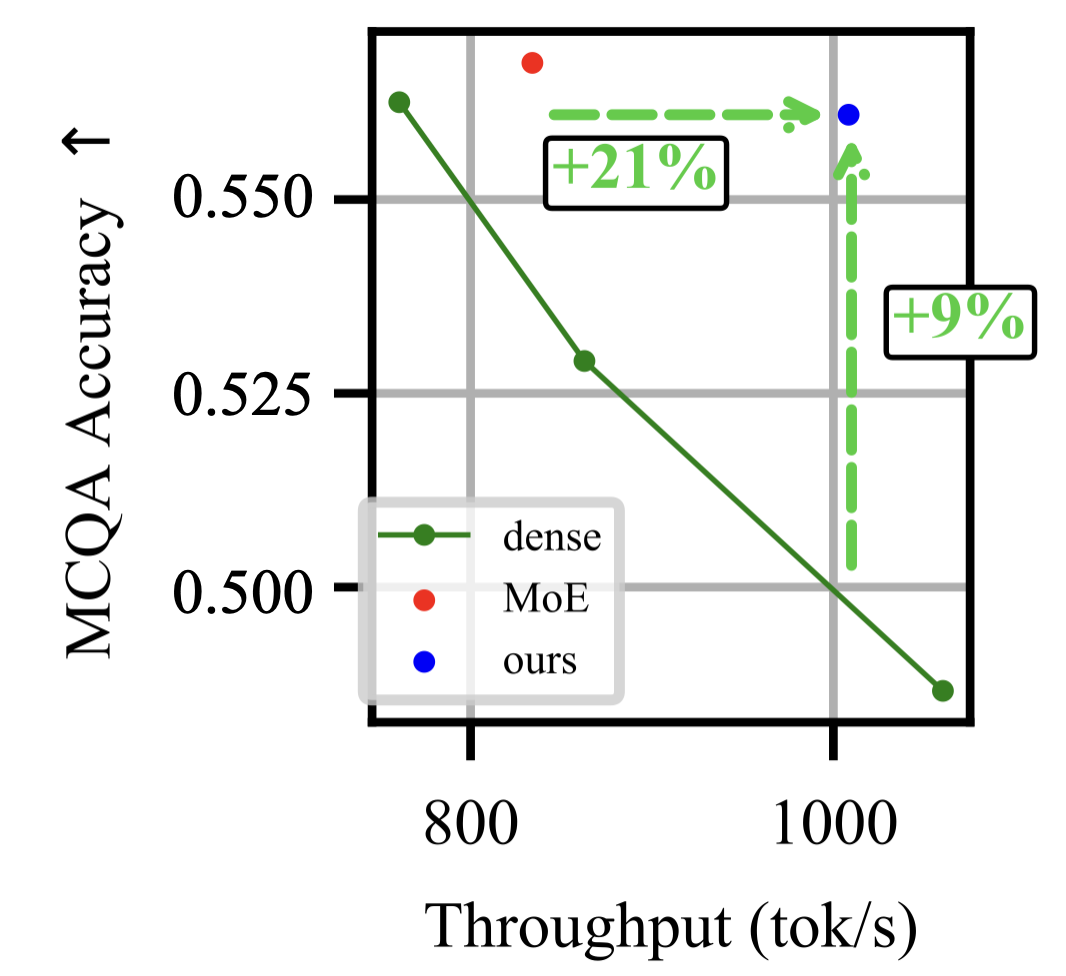}
    \caption{Accuracy}
    \label{fig:acc-vs-throughput}
\end{subfigure}
\caption{An overview of our method. (a): Standard MoEs require transferring many experts to the Neural Processing Unit (NPU) during SD, creating a bottleneck. (b): Our MoE improves expert coactivation, which increases throughput. (c): Our method improves throughput across all SD verification window sizes. (d): Our method retains accuracy benefits of MoE while improving throughput.}
\label{fig:overview}
\end{figure}
\section{Introduction}

Mixture-of-Experts (MoE) models have demonstrated strong results in language-modeling tasks \citep{Jacobs1991AdaptiveMO,Muennighoff2024OLMoEOM,Shazeer2017OutrageouslyLN,Zhou2022MixtureofExpertsWE}. Their appeal lies in their ability to scale up model sizes while reducing training and inference cost by activating only a subset of the model's feedforward network (FFN) parameters during the forward pass. This technique has become widely adopted among today's leading open-weight LLMs \citep{deepseekai2025deepseekv3technicalreport,Muennighoff2024OLMoEOM,Yang2025Qwen3,OpenAI2025GPTOSS,NVIDIA2026Nemotron3Super,Xiaomi2026MiMoV2Flash}.

In a transformer, MoE and dense models differ primarily in the FFN, where an MoE dynamically routes each token to a subset of expert FFNs (see \autoref{sec:background} for a formal treatment). This lets MoE LLMs match the latency and FLOPs of a much smaller dense model while approaching the quality of a much larger one.

However, deploying MoEs for inference presents challenges. Efficient inference often relies on Speculative Decoding (SD) \citep{Leviathan2022FastIF}, which produces multiple tokens per step using a cheap approximation model to draft tokens that the target model verifies in parallel (see \autoref{sec:background} for details). Indeed, several of the newest frontier MoEs now ship native speculative decoding support built into the MoE architecture itself \citep{NVIDIA2026Nemotron3Super,Xiaomi2026MiMoV2Flash}, underscoring the practical importance of the problem we tackle in this work.

When performing inference with an MoE, the verification cost is no longer fixed. In this case, a parallel decoding call will take significantly longer than a single-token decoding call. This is because neighboring tokens may choose different experts, which means that a greater number of expert weights need to be transferred from dynamic random access memory (DRAM) to the Neural Processing Unit, or NPU (e.g. a TPU, GPU, or other accelerator) compared to single-token inference. Since decoding is bottlenecked by model weight transfers during inference, this extra parameter transfer will result in a significant decrease in inference throughput. This will reduce the computational savings normally provided by using sparsely-activated MoEs. In the worst case, all $N$ experts could be transferred to the NPU. Concurrent work has begun to characterize this interaction between SD and MoE routing from a serving-time perspective \citep{Saxena2025Cascade,Huang2025MoESD}; in this work, we instead ask whether the MoE itself can be trained to make SD verification cheaper.

Our goal is to improve \emph{expert coactivation} -- e.g. to encourage neighboring tokens to choose similar sets of experts during verification to recover the performance benefits of SD. Rather than intervening at serving time, we achieve this entirely by changing how the MoE is \emph{trained}: our method is simple, and applies four changes to the standard MoE training setup, leaving the SD algorithm and the deployed drafting mechanism untouched. First, we replace the standard \emph{local} load-balancing loss with a global one \citep{Qiu2025DemonsIT}. Second, we use shared experts \citep{deepseekai2025deepseekv3technicalreport}. Third, we add a novel \emph{consistency loss} to encourage neighboring token's router logits to be similar. Finally, we add a novel \emph{autoregressive expert selection} mechanism to ensure expert selections are similar for neighboring tokens.

\autoref{fig:overview} shows an overview of our approach. Figure 1a shows a standard MoE router verifying four tokens, requiring nine unique experts to be transferred from DRAM to the NPU; Figure 1b shows the same verification under our coactivation techniques, reducing this to five experts. Figure 1c shows that our method improves throughput across all SD window sizes $\gamma$. Figure 1d shows results for an MoE with $1$ billion total and $353$ million active parameters, compared to a $352$ million parameter dense model: our method retains the accuracy benefits of MoEs ($9\%$ relative improvement over dense) while achieving $21\%$ higher throughput than standard MoEs. Our contributions are as follows:
\begin{itemize}[leftmargin=*,itemsep=0pt,parsep=0pt,topsep=2pt,partopsep=0pt]
    \item We analyze the performance of MoEs with SD and show that expert transfers decrease throughput.
    \item We show that improving expert coactivation decreases memory transfer costs and improves SD's efficacy for MoEs.
    \item We propose a combination of training-time techniques -- a global load-balancing loss, shared experts, a novel consistency loss, and a novel autoregressive expert selection -- that improve expert coactivation during inference with SD, thereby improving overall throughput.
\end{itemize}
\section{Background} \label{sec:background}
We provide a brief overview of MoEs and of SD in the following sections.
\subsection{Mixture-of-Experts}
MoEs are a more compute-efficient alternative to dense transformers \citep{Vaswani2017AttentionIA}, applying structured sparsity dynamically at inference time.

A standard MoE replaces the FFN with several \emph{expert} FFNs; a \emph{router} $R$ computes per-expert weights for each token, giving an FFN output of
\begin{align}
    \text{FFN}_{\text{MoE}}(x) &= \sum _{i=0} ^{N-1} R(i)E_i(x),
\end{align}
where $x$ is the input token, $N$ is the number of experts, $E_i$ is the $i^{th}$ expert, and $R(i)$ is its router weight. $R$ includes a TopK function ($R(i)>0$ for $k$ values of $i$), so $E_i(x)$ need only be computed for $k$ unique experts.

\subsection{Speculative Decoding}
Speculative Decoding (SD) is a general technique for improving inference speed. It uses a smaller \emph{approximation model} alongside the deployed \emph{target model}: the approximation model autoregressively produces $\gamma$ candidate tokens, and the target model verifies all $\gamma+1$ tokens (the input plus the $\gamma$ candidates) in a single parallel call, accepting or rejecting each candidate such that the output distribution exactly matches autoregressive decoding with the target model alone. We refer readers to \citet{Leviathan2022FastIF} for the full verification procedure.

The key insight leveraged by SD is that model inference for modest context lengths is bottlenecked by memory bandwidth between DRAM and the NPU. For this reason, a parallel decoding step with a modest window size (for instance, $8$) takes approximately the same amount of time as a decoding step with a window size of $1$, since the same number of model weights are transferred to the NPU in both cases. By speculating future tokens with a computationally inexpensive approximation model, SD allows a target model to produce multiple tokens per inference call, resulting in higher throughput. The improvement in throughput depends on the acceptance rate, or the probability of accepting tokens produced by the approximation model. The overall throughput can be approximated as
\begin{align}
    s = \dfrac{v(\gamma, \mathcal{A}, \mathcal{M})}{\mathcal{A}(1) \gamma + \mathcal{M}(\gamma + 1)},
\end{align}
where $v(\cdot)$ is the (average) number of tokens produced in an SD step, $\mathcal{A}$/$\mathcal{M}$ are the approximation/target models, and $\mathcal{A}(x)$/$\mathcal{M}(x)$ are their runtimes on $x$ input tokens. Note that $v(\cdot) \le \gamma + 1$, since at most $\gamma + 1$ tokens can be produced in one step. Throughput is measured in tokens generated per second.

\section{Motivation: Reduced Performance of SD on MoEs}
Our goal is to improve the inference-time throughput of MoEs used with SD. SD greatly improves throughput for dense models. However, we observe that MoEs benefit less from SD. This is because the SD verification step for MoEs requires transferring more experts than single-token inference. In other words, the standard assumption made by SD (i.e. that parallel and single-token inference take the same amount of time) is violated. To illustrate this point, we make the following observations:

\begin{table}
\centering
    \begin{tabular}{c||c | c | c | c }
      Model & message drafting & message rewriting & tone adjustment & machine translation \\
      \hline
      Dense 352M & $4.45\times$ &$2.54\times$ & $1.81\times$ & $3.50\times$ \\
      MoE X=16 K=4 & $3.38\times$ & $1.93\times$ & $1.42\times$ & $2.65\times$ \\
    \end{tabular}
\caption{Throughput improvements provided by SD for a dense 352M parameter model, and an MoE with 353M active parameters and 1032M total parameters (with 16 experts and topk set to 4).
    }
    \label{table:specd-improvements}
\end{table}

\begin{figure}
\centering
\begin{subfigure}[t]{0.31\textwidth}
    \centering
    \includegraphics[width=\textwidth]{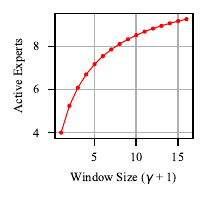}
    \caption{Experts vs Window Size}
    \label{fig:experts-vs-window}
\end{subfigure}
\begin{subfigure}[t]{0.31\textwidth}
    \centering
    \includegraphics[width=\textwidth]{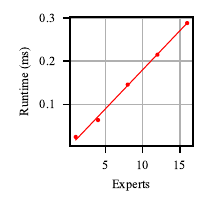}
    \caption{Runtime vs Experts}
    \label{fig:runtime-vs-experts}
\end{subfigure}
\begin{subfigure}[t]{0.31\textwidth}
    \centering
    \includegraphics[width=\textwidth]{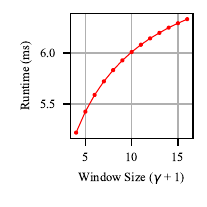}
    \caption{Throughput vs Window Size}
    \label{fig:runtime-vs-window}
\end{subfigure}

\caption{Larger window sizes increase the runtime of MoE FFN layers. (a): Larger window sizes increase the number of active experts. (b): Increasing the number of active experts increases the runtime. (c): The impact of window size on runtime.}
\label{fig:runtime-motivation}
\end{figure}

\textbf{Observation 1: SD gives strong throughput increase to dense models, but lower throughput increase for MoEs.} In \autoref{table:specd-improvements}, we view the improvements in throughput provided by SD for a dense model and an MoE for various tasks. Both models have the same number of active parameters, and the same single-token inference speed ($240$ tokens/s). However, the SD speedup for MoEs is reduced across the tasks. We investigate the source of this reduction in Observations 2 and 3.

\textbf{Observation 2: Parallel decoding increases the DRAM-to-NPU communication cost for MoE.} During parallel decoding, the number of experts activated within a layer of the MoE increases as the window size increases. In \autoref{fig:experts-vs-window}, we show the average number of active experts (per-layer) as a function of the SD verification window size (i.e. the number of tokens decoded in parallel by the target model). As the window size increases, the number of experts increases. This corresponds to an increase in the number of parameters that need to be sent to the NPU for computation.

\textbf{Observation 3: This increase in DRAM-to-NPU communication slows down the model.} The increase in the number of experts in Observation 2 increases inference cost. In \autoref{fig:runtime-vs-experts}, we benchmark the runtime of an MoE FFN layer as a function of the number of active experts. The trend is linear, with a non-zero y-intercept. As the number of experts increases, the runtime increases.

\textbf{Summary of Observations 1-3: Reduced performance of SD with MoEs is the result of increasing DRAM-to-NPU communication as window sizes increase.} In \autoref{fig:runtime-vs-window}, we visualize the runtime as a function of SD window size. For MoEs, the runtime increases steadily as window size increases. This reduces the speedup provided by SD, especially at high window sizes.

\section{Improving Expert Coactivation To Improve Speed}
To improve the runtime of MoEs with SD, we can reduce the number of experts activated by the set of tokens in the SD window. We hypothesize that we can encourage neighboring tokens to work together to choose overlapping experts with a minimal impact on accuracy. We accomplish this with four training-time changes to the MoE router: a global load-balancing loss (\autoref{sec:global-load-balancing-loss}), shared experts (\autoref{sec:shared-experts}), a novel consistency loss (\autoref{sec:consistency-loss}), and a novel autoregressive expert selection (\autoref{sec:autoregressive-expert-selection}). Finally, we demonstrate that these methods increase expert coactivation in \autoref{sec:coactivation}.

\subsection{Global Load-Balancing Loss} \label{sec:global-load-balancing-loss}
MoEs typically use a load-balancing loss to ensure that all experts are used by the network \citep{Shazeer2017OutrageouslyLN,Muennighoff2024OLMoEOM}. This helps prevent experts from becoming underutilized -- if an expert has a very low probability of being selected, the load balancing loss can help it recover.

The load balancing loss computed over a set of neighboring tokens $\tau$ is computed as

\begin{align}
    \mathcal{L}_{Bal}(\tau)=N \sum _{i=0} ^{N-1} f(i, \tau) p(i, \tau),
\end{align}
where $N$ is the number of experts, $f(i, \tau)$ is the fraction of tokens in $\tau$ that selected expert $i$, and $p$ is the total routing probability among tokens in $\tau$ routed to expert $i$.

On training clusters, processes across multiple nodes compute gradient updates in parallel, with the global batch $\mathcal{B}$ divided into micro-batches per process. Usually, the load-balancing loss is computed separately per micro-batch and averaged, to avoid inter-node data transfer \citep{Qiu2025DemonsIT}. The full loss across $\mathcal{B}$ is thus the average of local losses over each micro-batch of $\tau$ tokens,

\begin{align}
    \mathcal{L}_{Local}(\mathcal{B})&=\dfrac{|\tau|}{|\mathcal{B}|}\sum_{\tau \in \mathcal{B}}\mathcal{L}_{Bal}(\tau) \\
    &=\dfrac{|\tau|}{|\mathcal{B}|}\sum_{\tau \in \mathcal{B}} N \sum_{i=0} ^{N-1} f(i, \tau) p(i, \tau).
\end{align}

One drawback of this local loss is that it encourages the router to select all experts per micro-batch, \emph{even if the micro-batch's data come from the same domain}, adversely impacting expert specialization \citep{Qiu2025DemonsIT}. To alleviate this, we use a global load-balancing loss instead, formulated as

\begin{align}
    \mathcal{L}_{Global}(\mathcal{B}) &= \mathcal{L}_{Bal}(\mathcal{B}) \\
    &= N \sum _{i=0} ^{N-1} f(i, \mathcal{B}) p(i, \mathcal{B}).
\end{align}
This requires more inter-node communication than using local loss. However, we find in practice this step does not cause a bottleneck in training.

\subsection{Shared Experts} \label{sec:shared-experts}
One simple way to improve expert coactivation is to implement shared experts \citep{deepseekai2025deepseekv3technicalreport}. In this formulation, shared experts are activated by all tokens, whereas routed experts are selected per token via the router's top-$k$ mechanism. Given $N$ total experts, $S$ of which are shared, the MoE's FFN output can be computed as

\begin{align}
    \text{FFN}_{\text{MoE-Shared}}(x) = \sum_{i=0}^{S-1} E_i(x) + \sum_{i=S} ^{N-1}R(i)E_i(x).
\end{align}
This guarantees that at least $S$ experts will be shared among all tokens.

\subsection{Consistency Loss} \label{sec:consistency-loss}
We introduce a novel \emph{consistency loss} to encourage neighboring tokens to produce similar router logits. Specifically, we add a KL divergence loss between the router logits for token $i$ and $i+1$:

\begin{align*}
    \mathcal{L}_{C} &= KL(\text{SG}(R(i)) || R(i+1)),
\end{align*}
where $R(i)$ denotes the router logits for the $i^{th}$ token, KL denotes the Kullback-Leibler divergence \citep{Kullback1951OnInformation}, and SG denotes the stop-gradient operator. The consistency loss is multiplied by a scale factor $\lambda_C$.

\subsection{Autoregressive Expert Selection} \label{sec:autoregressive-expert-selection}
Finally, we introduce a novel autoregressive expert selection algorithm to replace the standard per-token TopK selection in the router $R$, under which neighboring tokens select experts independently. Within each window of $\gamma + 1$ tokens, we autoregressively choose experts for each token so that neighboring tokens' selections become correlated: each subsequent token retains at least $k-l$ experts chosen by the previous token, then fills the remaining selections with its own highest-ranked experts.

More formally, let $E_i$ denote the set of experts chosen by token $i$. Let $\text{top}(k, X, \tilde{R})$ denote the highest-ranked $k$ elements from the set $X$ using the corresponding router weights $\tilde{R}$ (i.e., the router weights prior to top-$k$ selection). Further, let $[N]$ denote the index set of all experts, $\{0, 1, ..., N-1\}$. Within an SD window of $\gamma+1$ tokens, we compute the selected experts for each token as

\begin{align}
    E_i = \begin{cases}
        & \text{top}(k, [N], \tilde{R}(i)) \text{ if } i=0 \\
        & \alpha_i \cup \text{top}(l, [N] \setminus \alpha_i, \tilde{R}(i)) \text{ if } i > 0,
    \end{cases}
\end{align}
where $\alpha_i = \text{top}(k-l, E_{i-1}, \tilde{R}(i))$ represents the top $k-l$ experts chosen by the \emph{previous} token, ranked by the \emph{current} token's router weights $\tilde{R}(i)$.

By construction, we retain at least $k-l$ experts from the previous token, and add at most $l$ new experts when selecting experts for the current token. Thus, for a window size of $\gamma + 1$ tokens, the number of experts processed is at most $k + \gamma l$.

\section{Results}

We demonstrate in \autoref{sec:coactivation} that our methods improve throughput, and here verify that they do not harm accuracy.

\begin{table}
\centering
    \begin{tabular}{c||c | c | c | c | c}
      Model & layers & model dim & FFN dim & Active Params (M) & Total Params (M) \\
      \hline
      Dense 352M & 24 & 1024 & 3072 & 352 & 352 \\ 
      \hline
      Dense 522M & 24 & 1024 & 5376 & 522 & 522 \\
      \hline
      Dense 692M & 24 & 1024 & 7680 & 692 & 692 \\
      \hline
      Dense 862M & 24 & 1024 & 9984 & 862 & 862 \\
      \hline
      Dense 1032M & 24 & 1024 & 12288 & 1032 & 1032 \\
      \hline
      \hline
      MoE X=16 K=2 & 24 & 1024 & 12288 & 240 & 1032 \\
      \hline
      MoE X=16 K=4 & 24 & 1024 & 12288 & 353 & 1032 \\
      \hline
      MoE X=16 K=6 & 24 & 1024 & 12288 & 466 & 1032 \\
    \end{tabular}
    \caption{A summary of models used in experiments.}
    \label{table:architectures}
\end{table}
\subsection{Experimental Details}
A summary of our transformer-based LLMs appears in \autoref{table:architectures}. Our smallest dense model has $352$M parameters, following \citet{Gunter2024AppleIF}; other dense variants scale up the FFN dimension, and our MoEs are obtained by converting that FFN to an MoE layer at each scale. Our approximation model used in SD is $3\%$ of the target MoE's size, incurring negligible overhead.

For the fairest possible comparison, we train all models ourselves from scratch. All hyperparameters including learning rate, batch size, number of iterations, loss function settings, etc. are identical for MoE and dense models. MoE training additionally includes a load-balancing loss and a router z-loss \citep{Zoph2022STMoE}, which have no dense-model counterpart. See \autoref{sec:hyperparameter-settings} for hyperparameter settings. Our models are trained on large-scale text data for 6 trillion tokens, the point at which the loss saturates.

\begin{figure}
\centering
\begin{subfigure}[t]{0.24\textwidth}
    \centering
    \includegraphics[width=\textwidth]{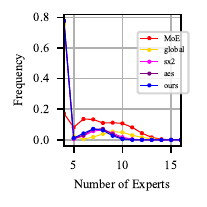}
    \caption{Layer 8}
    \label{fig:coactivation-local-global-8}
\end{subfigure}
\begin{subfigure}[t]{0.24\textwidth}
    \centering
    \includegraphics[width=\textwidth]{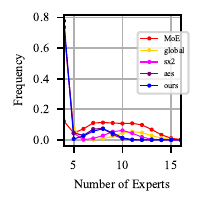}
    \caption{Layer 16}
    \label{fig:coactivation-local-global-16}
\end{subfigure}
\begin{subfigure}[t]{0.24\textwidth}
    \centering
    \includegraphics[width=\textwidth]{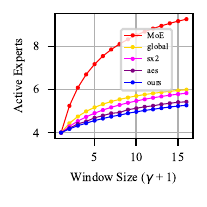}
    \caption{Active Experts}
    \label{fig:active-experts-window-size-both}
\end{subfigure}
\begin{subfigure}[t]{0.24\textwidth}
    \centering
    \includegraphics[width=\textwidth]{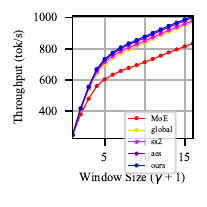}
    \caption{Throughput}
    \label{fig:throughput-gamma-teaser-2}
\end{subfigure}

\caption{Expert coactivation analysis for an MoE baseline with local load-balancing (\textbf{MoE}), global load-balancing (\textbf{global}), $2$ shared experts (\textbf{sx2}), autoregressive expert selection (\textbf{aes}), and all of the above with our auxiliary consistency loss (\textbf{ours}). (a) The frequency of active experts in an 8-token window at layer 8. (b) The frequency of active experts in an 8-token window at layer 16. (c) The average number of active experts as a function of window size. (d) Our improvements in coactivation translate to improvements in throughput at all SD window sizes.}
\label{fig:expert-coactivation}
\end{figure}

\subsection{Our Methods Improve Expert Coactivation} \label{sec:coactivation}
We investigate the improvements in expert coactivation using our four methods in \autoref{fig:expert-coactivation}. We ablate global load-balancing, shared experts, and autoregressive expert selection individually, since these directly control expert selection; our consistency loss is instead auxiliary, encouraging overlap only in the combined method (\textbf{ours}, in blue). For autoregressive expert selection, we choose $l=1$. Note that all methods use global load-balancing except the \textbf{MoE} baseline (in red), since local load-balancing harms expert specialization \citep{Qiu2025DemonsIT}. Our model has $N=16$ experts and $k=4$.

In \autoref{fig:coactivation-local-global-8}, we plot the total number of coactivated experts (x-axis) and the frequency (y-axis) for layer eight of the model when processing an eight-token window of the Wikipedia \citep{Gao2020ThePile} dataset. Switching from a local loss (\textbf{MoE}, red) to a global loss (\textbf{global}, yellow) drastically increases the frequency at which only four experts are selected across the eight-token window. We analyze layer sixteen of the model in \autoref{fig:coactivation-local-global-16} and see a similar pattern.

In \autoref{fig:active-experts-window-size-both}, we analyze the average number of active experts across a variety of window sizes. By combining all methods (\textbf{ours}, blue), we achieve the lowest average number of active experts. In \autoref{fig:throughput-gamma-teaser-2}, we observe that this translates to the strongest throughput at all window sizes.


\begin{figure}
\begin{subfigure}{0.24\textwidth}
    \centering
    \includegraphics[width=\textwidth]{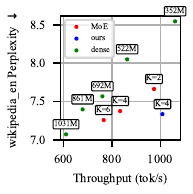}
    \caption{}
    \label{fig:acc-vs-throughput-wiki}
\end{subfigure}
\begin{subfigure}{0.24\textwidth}
    \centering
    \includegraphics[width=\textwidth]{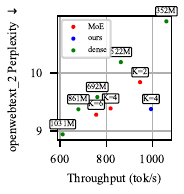}
    \caption{}
    \label{fig:acc-vs-throughput-owt2}
\end{subfigure}
\begin{subfigure}{0.24\textwidth}
    \centering
    \includegraphics[width=\textwidth]{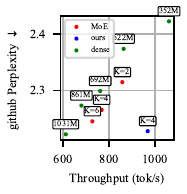}
    \caption{}
    \label{fig:acc-vs-throughput-github}
\end{subfigure}
\begin{subfigure}{0.24\textwidth}
    \centering
    \includegraphics[width=\textwidth]{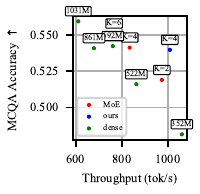}
    \caption{}
    \label{fig:acc-vs-throughput-qa}
\end{subfigure}
\caption{Perplexity (or accuracy) vs. throughput, comparing dense (green), MoE (red), and our MoE (blue), for (a) Wikipedia, (b) OpenWebText2, (c) Github, and (d) Question Answering. Our MoE produces the strongest results.}
\label{fig:efficiency-accuracy}
\end{figure}

\subsection{Our Method Outperforms Dense and MoE Baselines}
Here, we demonstrate that our method improves throughput over MoE baselines without adversely impacting accuracy. In \autoref{fig:efficiency-accuracy} we plot the throughput-accuracy trade-off curve of our method compared to dense and MoE baselines. We investigate the perplexity on Wikipedia \citep{Gao2020ThePile}, OpenWebText2 \citep{Gao2020ThePile}, and GitHub \citep{Gao2020ThePile}. We also investigate the accuracy of question-answering on a combination of Arc Easy \citep{Clark2018ARC}, Arc Challenge \citep{Clark2018ARC}, Hellaswag \citep{Zellers2019HellaSwag}, Lambada \citep{Paperno2016LAMBADA}, PiQA \citep{Bisk2020PIQA}, SciQ \citep{Welbl2017SciQ}, WinoGrande \citep{Sakaguchi2019WinoGrande}. In all cases, our model achieves a pareto-optimal throughput-accuracy trade-off. Our model is trained with $k=4$, and matches the accuracy of our MoE baseline with $k=4$; as shown in \autoref{fig:acc-vs-throughput-qa}, our MoE achieves this on-par accuracy while improving throughput by $21\%$ over the MoE baseline, due to increased coactivation. Our throughput surpasses that of an MoE baseline with only $k=2$ experts chosen.

\begin{figure}
\begin{subfigure}{0.24\textwidth}
    \centering
    \includegraphics[width=\textwidth]{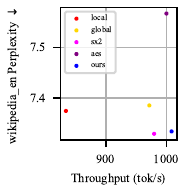}
    \caption{}
    \label{fig:acc-vs-throughput-wiki-all-methods}
\end{subfigure}
\begin{subfigure}{0.24\textwidth}
    \centering
    \includegraphics[width=\textwidth]{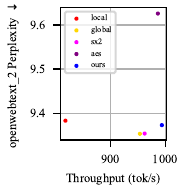}
    \caption{}
    \label{fig:acc-vs-throughput-owt2-all-methods}
\end{subfigure}
\begin{subfigure}{0.24\textwidth}
    \centering
    \includegraphics[width=\textwidth]{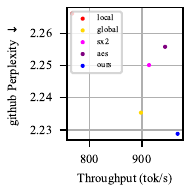}
    \caption{}
    \label{fig:acc-vs-throughput-github-all-methods}
\end{subfigure}
\begin{subfigure}{0.24\textwidth}
    \centering
    \includegraphics[width=\textwidth]{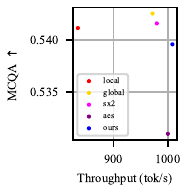}
    \caption{}
    \label{fig:acc-vs-throughput-qa-all-methods}
\end{subfigure}
\caption{Perplexity (or accuracy) vs. throughput, ablating local load-balancing (\textbf{local}), global load-balancing (\textbf{global}), $2$ shared experts (\textbf{sx2}), autoregressive expert selection (\textbf{aes}), and combining global, sx2, and aes with our auxiliary consistency loss (\textbf{ours}), for (a) Wikipedia, (b) OpenWebText2, (c) Github, and (d) Question Answering.}
\label{fig:efficiency-accuracy-all-methods}
\end{figure}

\subsection{Ablation: Combining All Four Methods Gives the Best Performance}
We ablate the throughput-accuracy trade-off provided by global load-balancing, shared experts, and autoregressive expert selection individually, versus combining them with our auxiliary consistency loss (\textbf{ours}). In \autoref{fig:acc-vs-throughput-qa-all-methods}, we compare our combined method (\textbf{ours}, blue) with applying each method separately. Autoregressive expert selection (\textbf{aes}, purple) yields a small reduction in quality, but performance recovers when combined with our other methods (\textbf{ours}, blue).

\begin{figure}
\begin{subfigure}{0.24\textwidth}
    \centering
    \includegraphics[width=\textwidth]{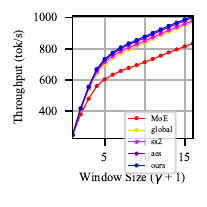}
    \caption{Message Drafting}
    \label{fig:throughput-vs-gamma-mail-reply}
\end{subfigure}
\begin{subfigure}{0.24\textwidth}
    \centering
    \includegraphics[width=\textwidth]{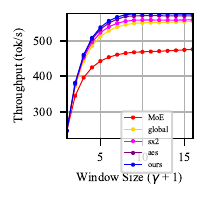}
    \caption{Message Rewriting}
    \label{fig:throughput-vs-gamma-magic-rewrite}
\end{subfigure}
\begin{subfigure}{0.24\textwidth}
    \centering
    \includegraphics[width=\textwidth]{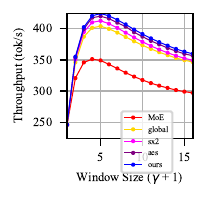}
    \caption{Tone Adjustment}
    \label{fig:throughput-vs-gamma-professional-tone}
\end{subfigure}
\begin{subfigure}{0.24\textwidth}
    \centering
    \includegraphics[width=\textwidth]{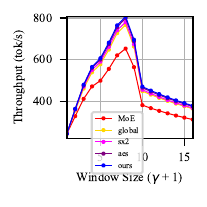}
    \caption{Machine Translation}
    \label{fig:throughput-vs-gamma-machine-translation}
\end{subfigure}
\caption{Throughput vs. SD window size for 4 different tasks.}
\label{fig:throughput-specd-tasks}
\end{figure}

\begin{figure}[h!]
\begin{subfigure}{0.24\textwidth}
    \centering
    \includegraphics[width=\textwidth]{figures/plots/mail_reply_wikipedia_en.pdf}
    \caption{Message Drafting}
    \label{fig:mail-reply-wikipedia}
\end{subfigure}
\begin{subfigure}{0.24\textwidth}
    \centering
    \includegraphics[width=\textwidth]{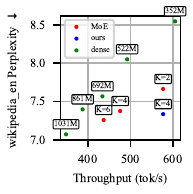}
    \caption{Message Rewriting}
    \label{fig:magic-rewrite-wikipedia}
\end{subfigure}
\begin{subfigure}{0.24\textwidth}
    \centering
    \includegraphics[width=\textwidth]{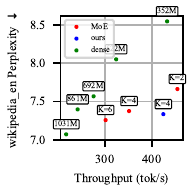}
    \caption{Tone Adjustment}
    \label{fig:professional-tone-wikipedia}
\end{subfigure}
\begin{subfigure}{0.24\textwidth}
    \centering
    \includegraphics[width=\textwidth]{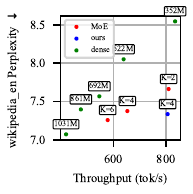}
    \caption{Machine Translation}
    \label{fig:machine-translation-wikipedia}
\end{subfigure}
\caption{Perplexity vs. throughput for acceptance rates corresponding to 4 different tasks.}
\label{fig:efficiency-accuracy-specd-tasks}
\end{figure}
\subsection{Impact of Task Difficulty}
Some tasks yield a higher SD acceptance rate than others: as seen in \autoref{table:specd-improvements}, \emph{message drafting} is easier for SD, whereas \emph{tone adjustment} is harder. We verify that our improvements translate across tasks. In \autoref{fig:throughput-specd-tasks}, we plot throughput vs. window size using SD acceptance rates for four tasks: our method has the highest peak throughput across all tasks, peaking at higher window sizes than baselines since coactivation is stronger.

For easier tasks (e.g., \emph{message drafting}), our method and baselines peak at higher window sizes, since larger windows are more likely to have more tokens accepted; for harder tasks, both peak at lower window sizes. In all cases, our model improves peak throughput over baselines.

In \autoref{fig:efficiency-accuracy-specd-tasks}, we plot the throughput-accuracy trade-off curve of our method compared to dense and MoE baselines using SD acceptance rates for these tasks. Across all tasks, our method achieves a stronger trade-off. For easier tasks (e.g. \emph{message drafting}), our method's advantage is stronger.

\section{Related Work}

MoEs scale model capacity while controlling inference cost by activating a sparse subset of expert FFNs per token \citep{Jacobs1991AdaptiveMO,Shazeer2017OutrageouslyLN,Zhou2022MixtureofExpertsWE}. \citet{Krajewski2024ScalingLaws} and \citet{Ludziejewski2025JointScaling} derive MoE scaling laws showing the compute-optimal expert count stays low until model scale exceeds several billion parameters. Our setting differs: we assume enough memory for all experts, with DRAM-to-NPU \emph{bandwidth} as the binding constraint, not compute. Local vs. global load-balancing losses also affect expert specialization \citep{Qiu2025DemonsIT}, which we build on in \autoref{sec:global-load-balancing-loss}. \citet{Skliar2024MixtureOC} explore post-training cache-reuse techniques for loading experts from flash on mobile devices; we instead target the number of \emph{distinct} experts requested per verification step. Instruction-conditioned pruning \citep{Hou2025InstructionPruning} offers an MoE-free route to sparsity, without retaining a single model across tasks.

Speculative Decoding \citep{Leviathan2022FastIF} was originally analyzed for dense models, where verification is free regardless of window size. Two recent studies show how this breaks down for MoEs: \citet{Saxena2025Cascade} show speculation can inflate expert-weight transfers by $2$-$3\times$, sometimes causing a net slowdown, and propose an adaptive policy; \citet{Huang2025MoESD} instead find MoEs can benefit \emph{more} from SD under the right batch-size regime. Other systems work reduces verification-time expert loads: training-free per-layer expert budgeting \citep{McDanel2026MoESpec}, cost-aware draft selection that avoids ``expert scattering'' \citep{Xie2026EcoSpec}, and commitment-weighted, self-sizing verifier expert sets \citep{Liang2026AcceptMoE}. All of these act at inference time on a fixed, trained MoE; our approach instead modifies router training, and is complementary to all of them.

Closest to our approach, \citet{Suram2026CacheableByDesign} train small ($\leq$340M-parameter) MoE routers for cache locality via auxiliary losses, but report a negative result: cache-miss reductions violate a pre-registered perplexity gate. \citet{Wu2026CacheAwareRouter} instead adapt the MoE backbone with an auxiliary cache-prediction module, improving cache hit rates without retraining router selection. We differ in objective and scale: we target \emph{coactivation within an SD verification window} rather than cache locality generally, evaluating at the billion-parameter scale, above where \citet{Suram2026CacheableByDesign} observe the tradeoff.

A separate line of work trains the MoE itself to draft for itself, e.g.\ via self-distillation of a per-layer draft expert \citep{Han2026DraftExpert} or self-speculative decoding on-device \citep{Huang2026S2MoE}. These methods modify SD's drafting side, whereas ours modifies only the target model's router, and could be combined with any of them.

\section{Conclusion}
We investigate the performance of MoEs with SD. We find that SD demonstrates reduced efficacy on MoEs due to increased DRAM-to-NPU bandwidth usage during verification. We show that applying global load-balancing, shared experts, autoregressive expert selection, and consistency loss during MoE training encourages expert coactivation, improving throughput on downstream tasks without harming accuracy. These improvements hold across a variety of tasks and datasets.

\subsection*{AI use statement}

We used generative AI tools for language editing: correcting grammar and reworking wordy
paragraphs into more compact prose. We did not use generative AI to conduct research, generate
research ideas, or draft any section of the manuscript from scratch; all text originates from the
authors. We have reviewed all AI-assisted edits and take responsibility for the final content of
this work, including text, claims, or artifacts produced with the aid of generative AI.

\bibliography{bibliography}

\begin{thebibliography}{35}
\providecommand{\natexlab}[1]{#1}
\providecommand{\url}[1]{\texttt{#1}}
\expandafter\ifx\csname urlstyle\endcsname\relax
  \providecommand{\doi}[1]{doi: #1}\else
  \providecommand{\doi}{doi: \begingroup \urlstyle{rm}\Url}\fi

\bibitem[Bisk et~al.(2019)Bisk, Zellers, Bras, Gao, and Choi]{Bisk2020PIQA}
Yonatan Bisk, Rowan Zellers, Ronan~Le Bras, Jianfeng Gao, and Yejin Choi.
\newblock Piqa: Reasoning about physical commonsense in natural language.
\newblock \emph{arXiv preprint arXiv:1911.11641}, 2019.
\newblock URL \url{https://arxiv.org/abs/1911.11641}.
\newblock AAAI 2020.

\bibitem[Clark et~al.(2018)Clark, Cowhey, Etzioni, Khot, Sabharwal, Schoenick,
  and Tafjord]{Clark2018ARC}
Peter Clark, Isaac Cowhey, Oren Etzioni, Tushar Khot, Ashish Sabharwal, Carissa
  Schoenick, and Oyvind Tafjord.
\newblock Think you have solved question answering? try arc, the ai2 reasoning
  challenge.
\newblock \emph{arXiv preprint arXiv:1803.05457}, 2018.
\newblock URL \url{https://arxiv.org/abs/1803.05457}.

\bibitem[DeepSeek-AI et~al.(2025)DeepSeek-AI, Liu, Feng, Xue, Wang, Wu, Lu,
  Zhao, Deng, Zhang, Ruan, Dai, Guo, Yang, Chen, Ji, Li, Lin, Dai, Luo, Hao,
  Chen, Li, Zhang, Bao, Xu, Wang, Zhang, Ding, Xin, Gao, Li, Qu, Cai, Liang,
  Guo, Ni, Li, Wang, Chen, Chen, Yuan, Qiu, Li, Song, Dong, Hu, Gao, Guan,
  Huang, Yu, Wang, Zhang, Xu, Xia, Zhao, Wang, Zhang, Li, Wang, Zhang, Zhang,
  Tang, Li, Tian, Huang, Wang, Zhang, Wang, Zhu, Chen, Du, Chen, Jin, Ge,
  Zhang, Pan, Wang, Xu, Zhang, Chen, Li, Lu, Zhou, Chen, Wu, Ye, Ye, Ma, Wang,
  Zhou, Yu, Zhou, Pan, Wang, Yun, Pei, Sun, Xiao, Zeng, Zhao, An, Liu, Liang,
  Gao, Yu, Zhang, Li, Jin, Wang, Bi, Liu, Wang, Shen, Chen, Zhang, Chen, Nie,
  Sun, Wang, Cheng, Liu, Xie, Liu, Yu, Song, Shan, Zhou, Yang, Li, Su, Lin, Li,
  Wang, Wei, Zhu, Zhang, Xu, Xu, Huang, Li, Zhao, Sun, Li, Wang, Yu, Zheng,
  Zhang, Shi, Xiong, He, Tang, Piao, Wang, Tan, Ma, Liu, Guo, Wu, Ou, Zhu,
  Wang, Gong, Zou, He, Zha, Xiong, Ma, Yan, Luo, You, Liu, Zhou, Wu, Ren, Ren,
  Sha, Fu, Xu, Huang, Zhang, Xie, Zhang, Hao, Gou, Ma, Yan, Shao, Xu, Wu,
  Zhang, Li, Gu, Zhu, Liu, Li, Xie, Song, Gao, and
  Pan]{deepseekai2025deepseekv3technicalreport}
DeepSeek-AI, Aixin Liu, Bei Feng, Bing Xue, Bingxuan Wang, Bochao Wu, Chengda
  Lu, Chenggang Zhao, Chengqi Deng, Chenyu Zhang, Chong Ruan, Damai Dai, Daya
  Guo, Dejian Yang, Deli Chen, Dongjie Ji, Erhang Li, Fangyun Lin, Fucong Dai,
  Fuli Luo, Guangbo Hao, Guanting Chen, Guowei Li, H.~Zhang, Han Bao, Hanwei
  Xu, Haocheng Wang, Haowei Zhang, Honghui Ding, Huajian Xin, Huazuo Gao, Hui
  Li, Hui Qu, J.~L. Cai, Jian Liang, Jianzhong Guo, Jiaqi Ni, Jiashi Li, Jiawei
  Wang, Jin Chen, Jingchang Chen, Jingyang Yuan, Junjie Qiu, Junlong Li,
  Junxiao Song, Kai Dong, Kai Hu, Kaige Gao, Kang Guan, Kexin Huang, Kuai Yu,
  Lean Wang, Lecong Zhang, Lei Xu, Leyi Xia, Liang Zhao, Litong Wang, Liyue
  Zhang, Meng Li, Miaojun Wang, Mingchuan Zhang, Minghua Zhang, Minghui Tang,
  Mingming Li, Ning Tian, Panpan Huang, Peiyi Wang, Peng Zhang, Qiancheng Wang,
  Qihao Zhu, Qinyu Chen, Qiushi Du, R.~J. Chen, R.~L. Jin, Ruiqi Ge, Ruisong
  Zhang, Ruizhe Pan, Runji Wang, Runxin Xu, Ruoyu Zhang, Ruyi Chen, S.~S. Li,
  Shanghao Lu, Shangyan Zhou, Shanhuang Chen, Shaoqing Wu, Shengfeng Ye,
  Shengfeng Ye, Shirong Ma, Shiyu Wang, Shuang Zhou, Shuiping Yu, Shunfeng
  Zhou, Shuting Pan, T.~Wang, Tao Yun, Tian Pei, Tianyu Sun, W.~L. Xiao,
  Wangding Zeng, Wanjia Zhao, Wei An, Wen Liu, Wenfeng Liang, Wenjun Gao,
  Wenqin Yu, Wentao Zhang, X.~Q. Li, Xiangyue Jin, Xianzu Wang, Xiao Bi,
  Xiaodong Liu, Xiaohan Wang, Xiaojin Shen, Xiaokang Chen, Xiaokang Zhang,
  Xiaosha Chen, Xiaotao Nie, Xiaowen Sun, Xiaoxiang Wang, Xin Cheng, Xin Liu,
  Xin Xie, Xingchao Liu, Xingkai Yu, Xinnan Song, Xinxia Shan, Xinyi Zhou,
  Xinyu Yang, Xinyuan Li, Xuecheng Su, Xuheng Lin, Y.~K. Li, Y.~Q. Wang, Y.~X.
  Wei, Y.~X. Zhu, Yang Zhang, Yanhong Xu, Yanhong Xu, Yanping Huang, Yao Li,
  Yao Zhao, Yaofeng Sun, Yaohui Li, Yaohui Wang, Yi~Yu, Yi~Zheng, Yichao Zhang,
  Yifan Shi, Yiliang Xiong, Ying He, Ying Tang, Yishi Piao, Yisong Wang, Yixuan
  Tan, Yiyang Ma, Yiyuan Liu, Yongqiang Guo, Yu~Wu, Yuan Ou, Yuchen Zhu, Yuduan
  Wang, Yue Gong, Yuheng Zou, Yujia He, Yukun Zha, Yunfan Xiong, Yunxian Ma,
  Yuting Yan, Yuxiang Luo, Yuxiang You, Yuxuan Liu, Yuyang Zhou, Z.~F. Wu,
  Z.~Z. Ren, Zehui Ren, Zhangli Sha, Zhe Fu, Zhean Xu, Zhen Huang, Zhen Zhang,
  Zhenda Xie, Zhengyan Zhang, Zhewen Hao, Zhibin Gou, Zhicheng Ma, Zhigang Yan,
  Zhihong Shao, Zhipeng Xu, Zhiyu Wu, Zhongyu Zhang, Zhuoshu Li, Zihui Gu,
  Zijia Zhu, Zijun Liu, Zilin Li, Ziwei Xie, Ziyang Song, Ziyi Gao, and Zizheng
  Pan.
\newblock Deepseek-v3 technical report, 2025.
\newblock URL \url{https://arxiv.org/abs/2412.19437}.

\bibitem[Gao et~al.(2020)Gao, Biderman, Black, Golding, Hoppe, Foster, Phang,
  He, Thite, Nabeshima, Presser, and Leahy]{Gao2020ThePile}
Leo Gao, Stella Biderman, Sid Black, Laurence Golding, Travis Hoppe, Charles
  Foster, Jason Phang, Horace He, Anish Thite, Noa Nabeshima, Shawn Presser,
  and Connor Leahy.
\newblock The pile: An 800gb dataset of diverse text for language modeling.
\newblock \emph{arXiv preprint arXiv:2101.00027}, 2020.

\bibitem[Gunter et~al.(2024)Gunter, Wang, Wang, Pang, Narayanan, Zhang, Zhang,
  Chen, Chiu, Qiu, Gopinath, Yap, Yin, Nan, Weers, Yin, Huang, Wang, Lu,
  Peebles, Ye, Lee, Du, Chen, Keunebroek, Wiseman, Evans, Lei, Rathod, Kong,
  Du, Li, Wang, Gao, Ahmed, Xu, Lu, Rashid, Jose, Doane, Bencomo, Vanderby,
  Hansen, Jain, Anupama, Kamal, Wu, Brum, Maalouf, Erdenebileg, Dulhanty,
  Moritz, Kang, Jimenez, Ladd, Shi, Bai, Chu, Hohman, Kotek, Coleman, Li,
  Bigham, Cao, Lai, Cheung, Shan, Zhou, Li, Qin, Singh, Vega, Zou, Heckman,
  Gardiner, Bowler, Cordell, Cao, Hay, Shahdadpuri, Godwin, Dighe, Rachapudi,
  Tantawi, Frigg, Davarnia, Shah, Guha, Sirovica, Ma, Ma, Wang, Kim, Jayaram,
  Shankar, Paidi, Kumar, Wang, Zheng, Cheng, Shrager, Ye, Tanaka, Guo, Meng,
  Luo, Zhi, Aygar, Wan, Walkingshaw, Lin, Farooq, Ramerth, Reed, Bartels,
  Chaney, Riazati, Yang, Feldman, Hochstrasser, Seguin, Belousova, Pelemans,
  Yang, Vahid, Cao, Najibi, Zuliani, Horton, Cho, Bhendawade, Dong, Maj,
  Agrawal, Shan, Fu, Poston, Xu, Liu, Rao, Heeramun, Merth, Rayala, Cui,
  Sridhar, Zhang, Zhang, Wu, Zhou, Liu, Zhao, Xia, Ren, and
  Ren]{Gunter2024AppleIF}
Tom Gunter, Zirui Wang, Chong Wang, Ruoming Pang, Andy Narayanan, Aonan Zhang,
  Bowen Zhang, Chen Chen, Chung-Cheng Chiu, David Qiu, Deepak Gopinath,
  Dian~Ang Yap, Dong Yin, Feng Nan, Floris Weers, Guoli Yin, Haoshuo Huang,
  Jianyu Wang, Jiarui Lu, John Peebles, Kewei Ye, Mark Lee, Nan Du, Qibin Chen,
  Quentin Keunebroek, Sam Wiseman, Syd Evans, Tao Lei, Vivek Rathod, Xiang
  Kong, Xianzhi Du, Yanghao Li, Yongqiang Wang, Yuan Gao, Zaid Ahmed, Zhaoyang
  Xu, Zhiyun Lu, Al~Rashid, Albin~Madappally Jose, Alec Doane, Alfredo Bencomo,
  Allison Vanderby, Andrew Hansen, Ankur Jain, Anupama~Mann Anupama, Areeba
  Kamal, Bugu Wu, Carolina Brum, Charlie Maalouf, Chinguun Erdenebileg, Chris
  Dulhanty, Dominik Moritz, Doug Kang, Eduardo Jimenez, Evan Ladd, Fang Shi,
  Felix Bai, Frank Chu, Fred Hohman, Hadas Kotek, Hannah~Gillis Coleman, Jane
  Li, Jeffrey~P. Bigham, Jeffery Cao, Jeff Lai, Jessica Cheung, Jiulong Shan,
  Joe Zhou, John Li, Jun Qin, Karanjeet Singh, Karla Vega, Kelvin Zou, Laura
  Heckman, Lauren Gardiner, Margit Bowler, Maria Cordell, Meng Cao, Nicole Hay,
  Nilesh Shahdadpuri, Otto Godwin, Pranay Dighe, Pushyami Rachapudi, Ramsey
  Tantawi, Roman Frigg, Sam Davarnia, Sanskruti Shah, Saptarshi Guha, Sasha
  Sirovica, Shen Ma, Shuang Ma, Simon Wang, Sulgi Kim, Suma Jayaram, Vaishaal
  Shankar, Varsha Paidi, Vivek Kumar, Xin Wang, Xin Zheng, Walker Cheng, Yael
  Shrager, Yang Ye, Yasu Tanaka, Yihao Guo, Yunsong Meng, Zhaoping Luo, Ouyang
  Zhi, Alp Aygar, Alvin Wan, Andrew~D. Walkingshaw, Tzu-Hsiang Lin, Arsalan
  Farooq, Brent Ramerth, Colorado Reed, Chris Bartels, Chris Chaney, David
  Riazati, Eric~Liang Yang, Erin Feldman, Gabriel Hochstrasser, Guillaume
  Seguin, Irina Belousova, Joris Pelemans, Karen Yang, Keivan~A. Vahid,
  Liangliang Cao, Mahyar Najibi, Marco Zuliani, Max Horton, Minsik Cho, Nikhil
  Bhendawade, Patrick Dong, Piotr Maj, Pulkit Agrawal, Qi~Shan, Qichen Fu,
  Regan Poston, Sam Xu, Shuangning Liu, Sushma Rao, Tashweena Heeramun, Thomas
  Merth, Uday Rayala, Victor Cui, Vivek~Rangarajan Sridhar, Wencong Zhang,
  Wenqi Zhang, Wentao Wu, Xingyu Zhou, Xinwen Liu, Yang Zhao, Yin Xia, Zhile
  Ren, and Zhongzheng Ren.
\newblock Apple intelligence foundation language models.
\newblock \emph{ArXiv}, abs/2407.21075, 2024.
\newblock URL \url{https://api.semanticscholar.org/CorpusID:271546289}.

\bibitem[Han(2026)]{Han2026DraftExpert}
Dengke Han.
\newblock Draftexpert: Expansion-aware self-speculative decoding for end-device
  moe inference, 2026.
\newblock URL \url{https://arxiv.org/abs/2607.24434}.

\bibitem[Hou et~al.(2025)Hou, Chen, Wang, Yin, Wang, Du, Pang, Chang, and
  Lei]{Hou2025InstructionPruning}
Bairu Hou, Qibin Chen, Jianyu Wang, Guoli Yin, Chong Wang, Nan Du, Ruoming
  Pang, Shiyu Chang, and Tao Lei.
\newblock Instruction-following pruning for large language models, 2025.
\newblock URL \url{https://arxiv.org/abs/2501.02086}.

\bibitem[Huang et~al.(2026)Huang, Qiu, and Li]{Huang2026S2MoE}
Haochen Huang, Shengxuan Qiu, and Meng Li.
\newblock S2-moe: Enabling efficient self-speculative decoding for
  mixture-of-experts on edge devices, 2026.
\newblock URL \url{https://arxiv.org/abs/2608.15018}.

\bibitem[Huang et~al.(2025)Huang, Zhu, Zhan, Hu, Mao, Yu, Liu, and
  Zhang]{Huang2025MoESD}
Zongle Huang, Lei Zhu, Zongyuan Zhan, Ting Hu, Weikai Mao, Xianzhi Yu, Yongpan
  Liu, and Tianyu Zhang.
\newblock Moesd: Unveil speculative decoding's potential for accelerating
  sparse moe, 2025.
\newblock URL \url{https://arxiv.org/abs/2505.19645}.
\newblock NeurIPS 2025 (spotlight).

\bibitem[Jacobs et~al.(1991)Jacobs, Jordan, Nowlan, and
  Hinton]{Jacobs1991AdaptiveMO}
Robert~A. Jacobs, Michael~I. Jordan, Steven~J. Nowlan, and Geoffrey~E. Hinton.
\newblock Adaptive mixtures of local experts.
\newblock \emph{Neural Computation}, 3:\penalty0 79--87, 1991.
\newblock URL \url{https://api.semanticscholar.org/CorpusID:572361}.

\bibitem[Krajewski et~al.(2024)Krajewski, Ludziejewski, Adamczewski, Pi{\'o}ro,
  Krutul, Antoniak, Ciebiera, Kr{\'o}l, Odrzyg{\'o}{\'z}d{\'z}, Sankowski,
  Cygan, and Jaszczur]{Krajewski2024ScalingLaws}
Jakub Krajewski, Jan Ludziejewski, Kamil Adamczewski, Maciej Pi{\'o}ro,
  Micha{\l} Krutul, Szymon Antoniak, Kamil Ciebiera, Krystian Kr{\'o}l, Tomasz
  Odrzyg{\'o}{\'z}d{\'z}, Piotr Sankowski, Marek Cygan, and Sebastian Jaszczur.
\newblock Scaling laws for fine-grained mixture of experts, 2024.
\newblock URL \url{https://arxiv.org/abs/2402.07871}.

\bibitem[Kullback \& Leibler(1951)Kullback and
  Leibler]{Kullback1951OnInformation}
Solomon Kullback and Richard~A Leibler.
\newblock On information and sufficiency.
\newblock \emph{The Annals of Mathematical Statistics}, 22\penalty0
  (1):\penalty0 79--86, 1951.

\bibitem[Leviathan et~al.(2022)Leviathan, Kalman, and
  Matias]{Leviathan2022FastIF}
Yaniv Leviathan, Matan Kalman, and Yossi Matias.
\newblock Fast inference from transformers via speculative decoding.
\newblock In \emph{International Conference on Machine Learning}, 2022.
\newblock URL \url{https://api.semanticscholar.org/CorpusID:254096365}.

\bibitem[Liang et~al.(2026)Liang, Chen, Mo, Wang, Li, Ma, and
  Luk]{Liang2026AcceptMoE}
Shuang Liang, Hao~Mark Chen, Zhiwen Mo, Qianzhou Wang, Guoyu Li, Lingxiao Ma,
  and Wayne Luk.
\newblock Acceptmoe: Commitment-weighted self-sizing verifier expert sets for
  efficient moe speculative decoding, 2026.
\newblock URL \url{https://arxiv.org/abs/2608.02989}.

\bibitem[Ludziejewski et~al.(2025)Ludziejewski, Pi{\'o}ro, Krajewski,
  Stefaniak, Krutul, Ma{\l}a{\'s}nicki, Cygan, Sankowski, Adamczewski,
  Mi{\l}o{\'s}, and Jaszczur]{Ludziejewski2025JointScaling}
Jan Ludziejewski, Maciej Pi{\'o}ro, Jakub Krajewski, Maciej Stefaniak,
  Micha{\l} Krutul, Jan Ma{\l}a{\'s}nicki, Marek Cygan, Piotr Sankowski, Kamil
  Adamczewski, Piotr Mi{\l}o{\'s}, and Sebastian Jaszczur.
\newblock Joint moe scaling laws: Mixture of experts can be memory efficient,
  2025.
\newblock URL \url{https://arxiv.org/abs/2502.05172}.

\bibitem[McDanel et~al.(2026)McDanel, Li, Surineni, and
  Khaitan]{McDanel2026MoESpec}
Bradley McDanel, Steven Li, Sruthikesh Surineni, and Harshit Khaitan.
\newblock Moe-spec: Expert budgeting for efficient speculative decoding, 2026.
\newblock URL \url{https://arxiv.org/abs/2602.16052}.

\bibitem[Muennighoff et~al.(2024)Muennighoff, Soldaini, Groeneveld, Lo,
  Morrison, Min, Shi, Walsh, Tafjord, Lambert, Gu, Arora, Bhagia, Schwenk,
  Wadden, Wettig, Hui, Dettmers, Kiela, Farhadi, Smith, Koh, Singh, and
  Hajishirzi]{Muennighoff2024OLMoEOM}
Niklas Muennighoff, Luca Soldaini, Dirk Groeneveld, Kyle Lo, Jacob~Daniel
  Morrison, Sewon Min, Weijia Shi, Pete Walsh, Oyvind Tafjord, Nathan Lambert,
  Yuling Gu, Shane Arora, Akshita Bhagia, Dustin Schwenk, David Wadden,
  Alexander Wettig, Binyuan Hui, Tim Dettmers, Douwe Kiela, Ali Farhadi,
  Noah~A. Smith, Pang~Wei Koh, Amanpreet Singh, and Hanna Hajishirzi.
\newblock Olmoe: Open mixture-of-experts language models.
\newblock \emph{ArXiv}, abs/2409.02060, 2024.
\newblock URL \url{https://api.semanticscholar.org/CorpusID:272366674}.

\bibitem[NVIDIA(2026)]{NVIDIA2026Nemotron3Super}
NVIDIA.
\newblock Nemotron 3 super: Open, efficient mixture-of-experts hybrid
  mamba-transformer model for agentic reasoning, 2026.
\newblock URL \url{https://arxiv.org/abs/2604.12374}.

\bibitem[OpenAI(2025)]{OpenAI2025GPTOSS}
OpenAI.
\newblock gpt-oss-120b \& gpt-oss-20b model card, 2025.
\newblock URL \url{https://arxiv.org/abs/2508.10925}.

\bibitem[Paperno et~al.(2016)Paperno, Kruszewski, Lazaridou, Pham, Bernardi,
  Pezzelle, Baroni, Boleda, and Fern{\'a}ndez]{Paperno2016LAMBADA}
Denis Paperno, Germ{\'a}n Kruszewski, Angeliki Lazaridou, Quan~Ngoc Pham,
  Raffaella Bernardi, Sandro Pezzelle, Marco Baroni, Gemma Boleda, and Raquel
  Fern{\'a}ndez.
\newblock The lambada dataset: Word prediction requiring a broad discourse
  context.
\newblock \emph{arXiv preprint arXiv:1606.06031}, 2016.
\newblock URL \url{https://arxiv.org/abs/1606.06031}.

\bibitem[Qiu et~al.(2025)Qiu, Huang, Zheng, Wen, Wang, Men, Titov, Liu, Zhou,
  and Lin]{Qiu2025DemonsIT}
Zihan Qiu, Zeyu Huang, Bo~Zheng, Kaiyue Wen, Zekun Wang, Rui Men, Ivan Titov,
  Dayiheng Liu, Jingren Zhou, and Junyang Lin.
\newblock Demons in the detail: On implementing load balancing loss for
  training specialized mixture-of-expert models.
\newblock \emph{ArXiv}, abs/2501.11873, 2025.
\newblock URL \url{https://api.semanticscholar.org/CorpusID:275787957}.

\bibitem[Sakaguchi et~al.(2019)Sakaguchi, Bras, Bhagavatula, and
  Choi]{Sakaguchi2019WinoGrande}
Keisuke Sakaguchi, Ronan~Le Bras, Chandra Bhagavatula, and Yejin Choi.
\newblock Winogrande: An adversarial winograd schema challenge at scale.
\newblock \emph{arXiv preprint arXiv:1907.10641}, 2019.
\newblock URL \url{https://arxiv.org/abs/1907.10641}.

\bibitem[Saxena et~al.(2025)Saxena, Tsai, Taneja, Jaleel, and
  Qureshi]{Saxena2025Cascade}
Anish Saxena, Po-An Tsai, Hritvik Taneja, Aamer Jaleel, and Moinuddin Qureshi.
\newblock Utility-driven speculative decoding for mixture-of-experts, 2025.
\newblock URL \url{https://arxiv.org/abs/2506.20675}.

\bibitem[Shazeer et~al.(2017)Shazeer, Mirhoseini, Maziarz, Davis, Le, Hinton,
  and Dean]{Shazeer2017OutrageouslyLN}
Noam~M. Shazeer, Azalia Mirhoseini, Krzysztof Maziarz, Andy Davis, Quoc~V. Le,
  Geoffrey~E. Hinton, and Jeff Dean.
\newblock Outrageously large neural networks: The sparsely-gated
  mixture-of-experts layer.
\newblock \emph{ArXiv}, abs/1701.06538, 2017.
\newblock URL \url{https://api.semanticscholar.org/CorpusID:12462234}.

\bibitem[Skliar et~al.(2024)Skliar, van Rozendaal, Lepert, Boinovski, van
  Baalen, Nagel, Whatmough, and Bejnordi]{Skliar2024MixtureOC}
Andrii Skliar, Ties van Rozendaal, Romain Lepert, Todor Boinovski, Mart van
  Baalen, Markus Nagel, Paul~N. Whatmough, and Babak~Ehteshami Bejnordi.
\newblock Mixture of cache-conditional experts for efficient mobile device
  inference.
\newblock \emph{ArXiv}, abs/2412.00099, 2024.
\newblock URL \url{https://api.semanticscholar.org/CorpusID:274436794}.

\bibitem[Suram(2026)]{Suram2026CacheableByDesign}
Shriniwas~Ramesh Suram.
\newblock Cacheable by design? training mixture-of-experts routers for locality
  against the edge memory-bandwidth wall, 2026.
\newblock URL \url{https://arxiv.org/abs/2608.18261}.

\bibitem[Vaswani et~al.(2017)Vaswani, Shazeer, Parmar, Uszkoreit, Jones, Gomez,
  Kaiser, and Polosukhin]{Vaswani2017AttentionIA}
Ashish Vaswani, Noam~M. Shazeer, Niki Parmar, Jakob Uszkoreit, Llion Jones,
  Aidan~N. Gomez, Lukasz Kaiser, and Illia Polosukhin.
\newblock Attention is all you need.
\newblock In \emph{Neural Information Processing Systems}, 2017.
\newblock URL \url{https://api.semanticscholar.org/CorpusID:13756489}.

\bibitem[Welbl et~al.(2017)Welbl, Liu, and Gardner]{Welbl2017SciQ}
Johannes Welbl, Nelson~F. Liu, and Matt Gardner.
\newblock Crowdsourcing multiple choice science questions.
\newblock \emph{arXiv preprint arXiv:1707.06209}, 2017.
\newblock URL \url{https://arxiv.org/abs/1707.06209}.

\bibitem[Wu et~al.(2026)Wu, Jin, Xing, Zhou, He, Wu, Cheng, Yang, and
  Chen]{Wu2026CacheAwareRouter}
Zhenhe Wu, Yaping Jin, Qinghua Xing, Hang Zhou, Wei He, Xianjie Wu, Xianfu
  Cheng, Jian Yang, and Hanting Chen.
\newblock Cache-aware joint router adaptation for memory-efficient moe
  inference, 2026.
\newblock URL \url{https://arxiv.org/abs/2609.04895}.

\bibitem[{Xiaomi LLM-Core Team}(2026)]{Xiaomi2026MiMoV2Flash}
{Xiaomi LLM-Core Team}.
\newblock Mimo-v2-flash technical report, 2026.
\newblock URL \url{https://arxiv.org/abs/2601.02780}.

\bibitem[Xie et~al.(2026)Xie, Liu, Huang, Ling, Dai, Zheng, and
  Hu]{Xie2026EcoSpec}
Jincheng Xie, Runheng Liu, Heyan Huang, Yawen Ling, Hanbin Dai, Yu~Zheng, and
  Wen Hu.
\newblock Less experts, faster decoding: Cost-aware speculative decoding for
  mixture-of-experts, 2026.
\newblock URL \url{https://arxiv.org/abs/2607.12696}.

\bibitem[Yang et~al.(2025)Yang, Li, Yang, Zhang, Hui, Zheng, Yu, Gao, Huang,
  Lv, Zheng, Liu, Zhou, Huang, Hu, Ge, Wei, Lin, Tang, Yang, Tu, Zhang, Yang,
  Yang, Zhou, Zhou, Lin, Dang, Bao, Yang, Yu, Deng, Li, Xue, Li, Zhang, Wang,
  Zhu, Men, Gao, Liu, Luo, Li, Tang, Yin, Ren, Wang, Zhang, Ren, Fan, Su,
  Zhang, Zhang, Wan, Liu, Wang, Cui, Zhang, Zhou, and Qiu]{Yang2025Qwen3}
An~Yang, Anfeng Li, Baosong Yang, Beichen Zhang, Binyuan Hui, Bo~Zheng, Bowen
  Yu, Chang Gao, Chengen Huang, Chenxu Lv, Chujie Zheng, Dayiheng Liu, Fan
  Zhou, Fei Huang, Feng Hu, Hao Ge, Haoran Wei, Huan Lin, Jialong Tang, Jian
  Yang, Jianhong Tu, Jianwei Zhang, Jianxin Yang, Jiaxi Yang, Jing Zhou,
  Jingren Zhou, Junyang Lin, Kai Dang, Keqin Bao, Kexin Yang, Le~Yu, Lianghao
  Deng, Mei Li, Mingfeng Xue, Mingze Li, Pei Zhang, Peng Wang, Qin Zhu, Rui
  Men, Ruize Gao, Shixuan Liu, Shuang Luo, Tianhao Li, Tianyi Tang, Wenbiao
  Yin, Xingzhang Ren, Xinyu Wang, Xinyu Zhang, Xuancheng Ren, Yang Fan, Yang
  Su, Yichang Zhang, Yinger Zhang, Yu~Wan, Yuqiong Liu, Zekun Wang, Zeyu Cui,
  Zhenru Zhang, Zhipeng Zhou, and Zihan Qiu.
\newblock Qwen3 technical report, 2025.
\newblock URL \url{https://arxiv.org/abs/2505.09388}.

\bibitem[Zellers et~al.(2019)Zellers, Holtzman, Bisk, Farhadi, and
  Choi]{Zellers2019HellaSwag}
Rowan Zellers, Ari Holtzman, Yonatan Bisk, Ali Farhadi, and Yejin Choi.
\newblock Hellaswag: Can a machine really finish your sentence?
\newblock \emph{arXiv preprint arXiv:1905.07830}, 2019.
\newblock URL \url{https://arxiv.org/abs/1905.07830}.

\bibitem[Zhou et~al.(2022)Zhou, Lei, Liu, Du, Huang, Zhao, Dai, Chen, Le, and
  Laudon]{Zhou2022MixtureofExpertsWE}
Yan-Quan Zhou, Tao Lei, Han-Chu Liu, Nan Du, Yanping Huang, Vincent Zhao,
  Andrew~M. Dai, Zhifeng Chen, Quoc~V. Le, and James Laudon.
\newblock Mixture-of-experts with expert choice routing.
\newblock \emph{ArXiv}, abs/2202.09368, 2022.
\newblock URL \url{https://api.semanticscholar.org/CorpusID:247011948}.

\bibitem[Zoph et~al.(2022)Zoph, Bello, Kumar, Du, Huang, Dean, Shazeer, and
  Fedus]{Zoph2022STMoE}
Barret Zoph, Irwan Bello, Sameer Kumar, Nan Du, Yanping Huang, Jeff Dean, Noam
  Shazeer, and William Fedus.
\newblock St-moe: Designing stable and transferable sparse expert models.
\newblock \emph{ArXiv}, abs/2202.08906, 2022.
\newblock URL \url{https://arxiv.org/abs/2202.08906}.

\end{thebibliography}
\bibliographystyle{iclr2027_conference}


\appendix
\section{Appendix}

\section{Hyperparameter Settings} \label{sec:hyperparameter-settings}
We train all models (dense and MoE) with the following settings.

\textbf{Training.} Batch size $512$ and sequence length $4096$. All models are trained on NVIDIA H100 GPUs.

\textbf{Optimization.} We use the AdaStar optimizer (EMA $=0.95$) with a learning rate of
$0.01$, linear warmup for $5000$ steps followed by cosine decay ($\alpha=0.005$), and weight
decay $3.16\times10^{-4}$. Gradients are rescaled to norm $1$ if their norm exceeds $1$, and
updates are skipped entirely if the norm exceeds $100$.

\textbf{Architecture.} Vocabulary size $49{,}152$, $16$ attention heads, and $8$ KV heads.

\textbf{Loss functions.} The dense model loss is cross-entropy plus a z-loss (scale
$10^{-6}$). The MoE loss additionally includes a load-balancing loss (weight $0.01$) and a
router z-loss (weight $10^{-4}$).


\end{document}